\documentclass[10pt,twocolumn,letterpaper]{article}

\usepackage{graphicx}
\usepackage{subcaption}
\usepackage{float}
\usepackage[justification=raggedright]{caption}	
\usepackage{lscape}                                         
\usepackage{wrapfig}
\usepackage[percent]{overpic}

\usepackage[lined,ruled,linesnumbered]{algorithm2e}
\usepackage{animate} 

\usepackage{booktabs}                   
\usepackage{multirow}

\usepackage{makecell}

\usepackage{paralist}
\usepackage{enumitem}

\usepackage{bm}                          
\usepackage{epsfig}                      
\usepackage{graphicx}                  
\usepackage{times}
\usepackage{mathtools}
\usepackage{amssymb,amsmath}   

\usepackage{units}
\usepackage{color}

\usepackage{comment}

\usepackage{url}  
\usepackage[pagebackref=true,breaklinks=true,letterpaper=true,colorlinks,bookmarks=false]{hyperref}
\usepackage{xspace}
\usepackage[table]{xcolor}
\usepackage{setspace}
\usepackage{grfext}
\PrependGraphicsExtensions*{.jpg,.png,.PNG}

\def\sfw{\mathbf{s}_\textit{fw}}
\def\sbw{\mathbf{s}_\textit{bw}}
\def\rij{\mathbf{r}_{ij}}

\newcommand{\norm}[1]{\left\lVert#1\right\rVert}
\def\x{\mb{x}} 
\def\d{\mb{d}} 
\def\c{\mb{c}} 
\def\cs{\mathbf{c}^s} 
\def\cd{\mathbf{c}^d} 
\def\interval{u} 
\def\r{\mathbf{r}} 
\def\s{\mb{s}} 

\newlength\paramargin
\newlength\figmargin

\newlength\secmargin
\newlength\figcapmargin
\newlength\tabcapmargin
\newlength\eqnmargin

\newcommand{\red}{\textcolor{red}}
\newcommand{\blue}{\textcolor{blue}}

\newcommand {\first}[1]{{\color{red}\textbf{#1}}}
\newcommand {\second}[1]{{\color{blue}\underline{#1}}}

\newcommand{\mpage}[2]
{
\begin{minipage}{#1\linewidth}\centering
#2
\end{minipage}
}

\newcommand{\topic}[1]
{
\vspace{1mm}\noindent\textbf{#1}
}

\newcommand{\secref}[1]{Section~\ref{sec:#1}}
\newcommand{\figref}[1]{Figure~\ref{fig:#1}} 
\newcommand{\tabref}[1]{Table~\ref{tab:#1}}

\long\def\ignorethis#1{}

\newcommand{\tb}[1]{\textbf{#1}}
\newcommand{\mb}[1]{\mathbf{#1}}

\newbox\jsavebox%

\def\xi{\mathbf{x}_i}

\graphicspath{{figure}, {example}}

\usepackage{iccv} 
\usepackage{animate}
\usepackage{subfiles} 
\def\etal{et al.}	

\iccvfinalcopy 
\def\iccvPaperID{2890} 

\ificcvfinal\fi
\begin{document}

\title{Dynamic View Synthesis from Dynamic Monocular Video}

\author{
Chen Gao\\
Virginia Tech\\
\and
Ayush Saraf\\
Facebook\\
\and
Johannes Kopf\\
Facebook\\
\and
Jia-Bin Huang\\
Virginia Tech\\
}


\twocolumn[{
\renewcommand\twocolumn[1][]{#1}
\maketitle
\begin{center}
\begin{overpic}[width=1\linewidth]{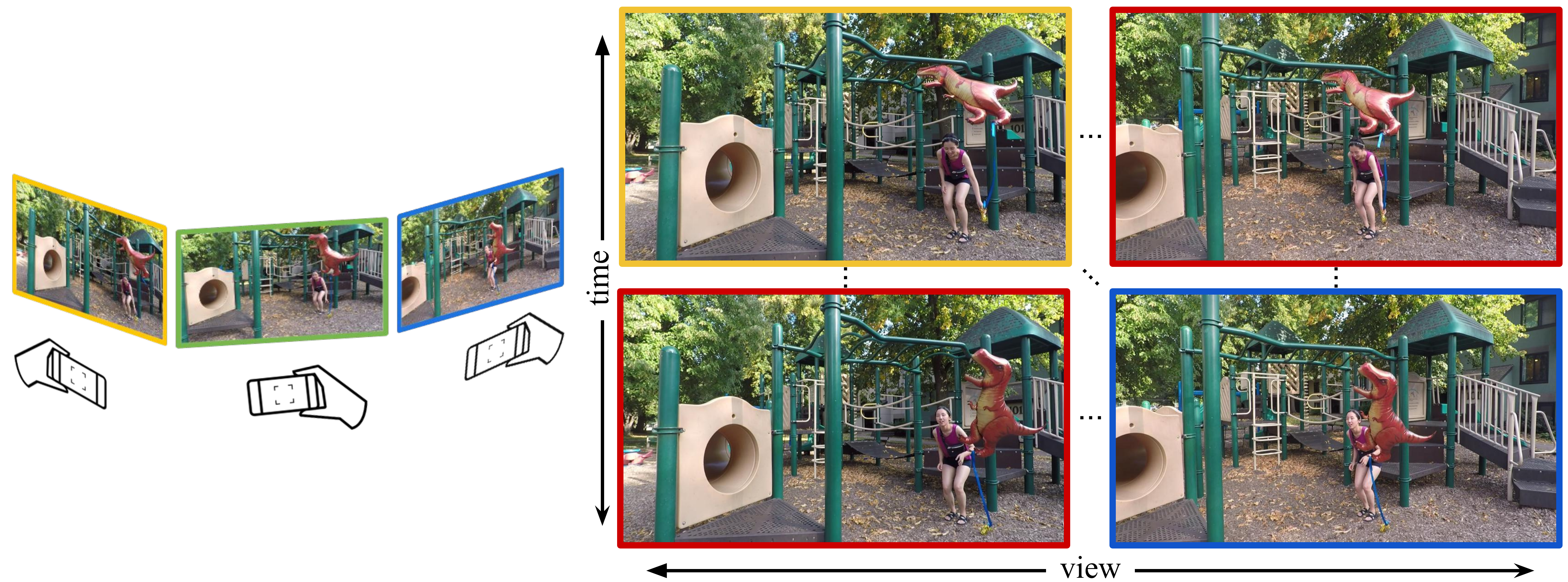}
\put (4, 8) {$t_\mathrm{0}$}
\put (18, 7) {$t_\mathrm{1}$}
\put (31, 9.5) {$t_\mathrm{N}$}
\put (25, 8.5) {\rotatebox[origin=c]{20}{$\dots$}}
\end{overpic}
\mpage{0.35}{(a) Input: monocular video}
\hfill
\mpage{0.6}{(b) Output: free-viewpoint rendering}
\vspace{\figcapmargin}
\captionof{figure}{
\tb{Dynamic view synthesis from dynamic monocular video.}
%
Our method takes a monocular video as input (a). 
Each frame in the video is taken at a unique time step and from a different view (e.g., the \textbf{\textcolor[HTML]{dfcf44}{yellow}} and \textbf{\blue{blue}} frames).
Our goal is to synthesize photorealistic novel views of a dynamic scene at arbitrary camera viewpoints and time steps (\textbf{\red{red}} frames).
Such a system enables \emph{free-viewpoint video}, providing immersive and almost life-like viewing experiences for users.
}
\end{center}
}]

\thispagestyle{empty}
\begin{abstract}
We present an algorithm for generating novel views at arbitrary viewpoints and any input time step given a monocular video of a dynamic scene.
Our work builds upon recent advances in neural implicit representation and uses continuous and differentiable functions for modeling the time-varying structure and the appearance of the scene.
We jointly train a time-invariant static NeRF and a time-varying dynamic NeRF, and learn how to blend the results in an unsupervised manner.
However, learning this implicit function from a single video is highly ill-posed (with infinitely many solutions that match the input video). 
To resolve the ambiguity, we introduce regularization losses to encourage a more physically plausible solution.
We show extensive quantitative and qualitative results of dynamic view synthesis from casually captured videos.
\end{abstract}
\section{Introduction}
\label{sec:intro}


Video provides a window into another part of the real world.
In traditional videos, however,  the viewer observes the action from a fixed viewpoint and cannot navigate the scene.
\emph{Dynamic view synthesis} comes to the rescue.
These techniques aim at creating photorealistic novel views of a dynamic scene at arbitrary camera viewpoints and time, which
enables \emph{free-viewpoint video} and \emph{stereo rendering}, and provides an immersive and almost life-like viewing experience.
It facilitates applications such as replaying professional sports events in 3D~\cite{Canon}, creating cinematic effects like freeze-frame bullet-time (from the movie ``The Matrix"), virtual reality~\cite{collet2015high,broxton2020immersive}, and virtual 3D teleportation~\cite{orts2016holoportation}.

Systems for dynamic view synthesis need to overcome challenging problems related to video capture, reconstruction, compression, and rendering. 
Most of the existing methods rely on laborious and expensive setups such as custom fixed multi-camera video capture rigs~\cite{carranza2003free,zitnick2004high,collet2015high,orts2016holoportation,broxton2020immersive}.
While recent work relaxes some constraints and can handle \emph{unstructured} video input (e.g., from hand-held cameras)~\cite{ballan2010unstructured,bansal20204d}, many methods still require synchronous capture from multiple cameras, which is impractical for most people.
Few methods produce dynamic view synthesis from a \emph{single} stereo or even RGB camera, but they are limited to specific domains such as human performance capture \cite{dou2016fusion4d,habermann2019livecap}.
Recent work on depth estimation from monocular videos of dynamic scenes shows promising results~\cite{Luo-VideoDepth-2020,Yoon-2020-CVPR}.
Yoon et al.~\cite{Yoon-2020-CVPR} use estimated depth maps to warp and blend multiple images to synthesize an unseen target viewpoint.
However, the method uses a \emph{local} representation (i.e., per-frame depth maps) and processes each novel view \emph{independently}. 
Consequently, the synthesized views are not consistent and may exhibit abrupt changes.

This paper presents a new algorithm for dynamic view synthesis from a dynamic video that overcomes this limitation using a \emph{global} representation. 
More specifically, we use an implicit neural representation to model the time-varying volume density and appearance of the events in the video.
We jointly train a time-invariant static neural radiance field (NeRF)~\cite{mildenhall2020nerf} and a time-varying dynamic NeRF, and learn how to blend the results in an unsupervised manner.
However, it is challenging for the dynamic NeRF to learn plausible 3D geometry because we have just \emph{one and only one} 2D image observation at each time step. 
There are infinitely many solutions that can correctly render the given input video, yet only one is physically correct for generating photorealistic novel views.
Our work focuses on resolving this ambiguity by introducing regularization losses to encourage plausible reconstruction.
We validate our method's performance on the Dynamic multi-view dynamic scenes dataset by Yoon et al.~\cite{Yoon-2020-CVPR}.

The key points of our contribution can be summarized as follows:
\begin{itemize}
\item We present a method for modeling dynamic radiance fields by jointly training a time-invariant model and a time-varying model, and learn how to blend the results in an unsupervised manner.
\item We design regularization losses for resolving the ambiguities when learning the dynamic radiance fields. 
\item Our model leads to favorable results compared to the state-of-the-art
algorithms on the Dynamic Scenes Dataset.
\end{itemize}

\section{Related Work}
\label{sec:related}






\topic{View synthesis from images.}
View synthesis aims to generate new views of a scene from multiple posed images \cite{shum2000review}.
Light fields \cite{levoy1996light} or Lumigraph~\cite{gortler1996lumigraph} synthesize realistic appearance but require capturing and storing many views.
Using explicit geometric proxies allows high-quality synthesis from relatively fewer input images \cite{buehler2001unstructured,Gao-portraitnerf}.
However, estimating accurate scene geometry is challenging due to untextured regions, highlights, reflections, and repetitive patterns.
Prior work addresses this via local warps \cite{chaurasia2013depth}, operating in the gradient domain~\cite{Kopf2013}, soft 3D reconstruction~\cite{penner2017soft}, and learning-based approaches~\cite{kalantari2016learning,flynn2016deepstereo,flynn2019deepview,hedman2018deep,riegler2020free}.
Recently, neural implicit representation methods have shown promising view synthesis results by modeling the continuous volumetric scene density and color with a multilayer perceptron~\cite{mildenhall2020nerf,Niemeyer2020DifferentiableVR,yariv2020multiview,zhang2020nerf++}.

Several methods tackle novel view synthesis from one single input image. 
These methods differ in their underlying scene representation, including depth~\cite{niklaus20193d,wiles2020synsin}, multiplane images~\cite{tucker2020single}, or layered depth images~\cite{shih20203d,kopf2020one}.
Compared with existing view synthesis methods that focus on \emph{static objects or scenes}, our work aims to achieve view synthesis of \emph{dynamic scenes} from one single video.

\topic{View synthesis for videos.}
Free viewpoint video offers immersive viewing experiences and creates freeze-frame (bullet time) visual effects \cite{magnor2005video}.
Compared to view synthesis techniques for images, capturing, reconstructing, compressing, and rendering dynamic contents in videos is significantly more challenging. 
Many existing methods either focus on specific domains (e.g., humans) \cite{carranza2003free,dou2016fusion4d,habermann2019livecap} or transitions between input views only~\cite{ballan2010unstructured}.
Several systems have been proposed to support interactive viewpoint control watching videos of generic scenes \cite{zitnick2004high,collet2015high,orts2016holoportation,bansal20204d,broxton2020immersive,attal2020matryodshka}.
However, these methods require either omnidirectional stereo camera~\cite{attal2020matryodshka}, specialized hardware setup (e.g., custom camera rigs) \cite{zitnick2004high,collet2015high,broxton2020immersive,orts2016holoportation}, or synchronous video captures from multiple cameras~\cite{bansal20204d}.
Recently, Yoon et al.~\cite{Yoon-2020-CVPR} show that one can leverage depth-based warping and blending techniques in image-based rendering for synthesizing novel views of a dynamic scene from a single camera.
Similar to \cite{Yoon-2020-CVPR}, our method also synthesizes novel views of a dynamic scene.
In contrast to using explicit depth estimation \cite{Yoon-2020-CVPR}, our implicit neural representation based approach facilitates geometrically accurate rendering and smoother view interpolation. 

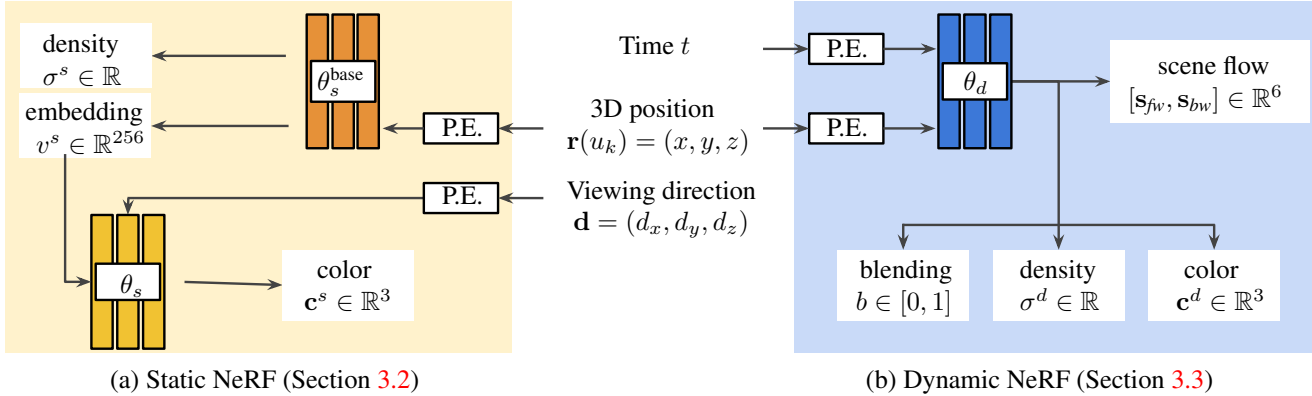
\begin{figure*}[t]
\begin{overpic}[width=1\linewidth]{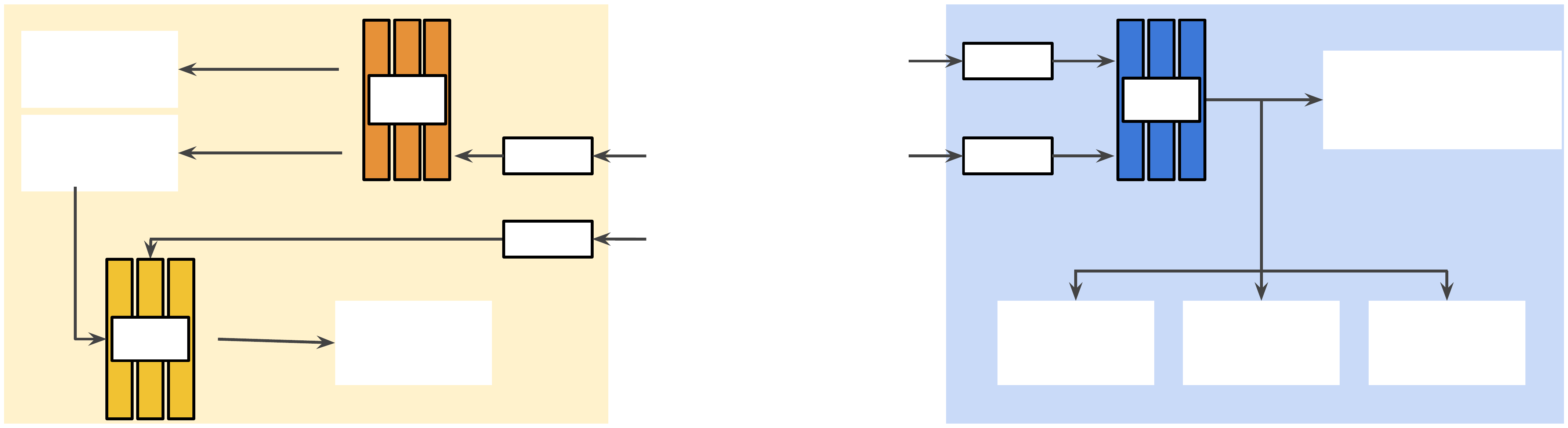}
\put (3.1, 23.2) {density}
\put (3.1, 20.7) {$\sigma^s \in \mathbb{R}$}
\put (1.8, 18) {embedding}
\put (2.6, 15.5) {$v^s \in \mathbb{R}^{256}$}
\put (24, 6) {color}
\put (23, 3.5) {$\cs \in \mathbb{R}^3$}
\put (88, 21.5) {scene flow}
\put (85.7, 19) {$\left[ \s_\textit{fw}, \s_\textit{bw} \right] \in \mathbb{R}^6$}
\put (65.3, 6) {blending}
\put (65.0, 3.5) {$b \in [0, 1]$}
\put (77.5, 6) {density}
\put (77.5, 3.5) {$\sigma^d \in \mathbb{R}$}
\put (90, 6) {color}
\put (89.4, 3.5) {$\cd \in \mathbb{R}^3$}
\put (73.2, 20.3) {$\theta_d$}
\put (9, 4.8) {$\theta_s$}
\put (24, 20.3) {$\theta^\text{base}_s$}
\put (44.8, 18.2) {3D position}
\put (43, 15.8) {$\r(\interval_k) = (x, y, z)$}
\put (47, 23) {Time $t$}
\put (43, 12) {Viewing direction}
\put (43.5, 9.6) {$\d = (d_x, d_y, d_z)$}

\put (33.5, 16.8) {P.E.}
\put (33.5, 11.6) {P.E.}
\put (62.8, 16.8) {P.E.}
\put (62.8, 22.8) {P.E.}
\end{overpic}
\mpage{0.4}{(a) Static NeRF (\secref{static_NeRF})} \hfill
\mpage{0.4}{(b) Dynamic NeRF (\secref{dynamic_NeRF})}
\vspace{\figcapmargin}
\caption{\textbf{Method overview.} 
We propose to use two different models to represent the (a) \emph{static} and (b) \emph{dynamic} scene components.
\tb{(a) Static NeRF}: For static components, we train a NeRF model following~\cite{mildenhall2020nerf}, but excluding all the pixels marked as dynamic from training the model.
This allows us to reconstruct the background's structure and appearance without conflicting the moving objects.
\tb{(b) Dynamic NeRF}: Modeling a dynamic scene from a single video is highly ill-posed. 
To resolve the ambiguity, we leverage the multi-view constraints as follow: 
Our Dynamic NeRF takes both $\r(\interval_k)$ and $t$ as input to predict 3D scene flow from time $t$ to $t+1$ ($\s_\textit{fw}$) and from time $t$ to $t-1$ ($\s_\textit{bw}$).
Using the predicted scene flow, we can create a \emph{warped} radiance field by resampling the radiance field modeled at the adjacent time instances and apply temporal consistency.
Thus, at each instance, we can have multiple views associated with different time $t$ to train the model. 
}
\label{fig:overview}
\end{figure*}


\topic{Implicit neural representations.}
Continuous and differentiable functions parameterized by fully-connected networks (also known as multilayer perceptron, or MLPs) have been successfully applied as compact, implicit representations for modeling 
3D shapes~\cite{chen2019learning,xu2019disn,park2019deepsdf,genova2019learning,genova2020local}, 
object appearances~\cite{oechsle2019texture,niemeyer2020differentiable},
3D scenes~\cite{sitzmann2019scene,mildenhall2020nerf,peng2020convolutional}.
These methods train MLPs to regress input coordinates (e.g., points in 3D space) to the desired quantities such as 
occupancy value~\cite{mescheder2019occupancy,saito2019pifu,peng2020convolutional}, 
signed distance~\cite{park2019deepsdf,atzmon2020sal,michalkiewicz2019implicit}, 
volume density~\cite{mildenhall2020nerf}, 
color~\cite{oechsle2019texture,sitzmann2019scene,saito2019pifu,mildenhall2020nerf}.
Leveraging differentiable rendering~\cite{mantiuk2020state,kato2020differentiable}, several recent works have shown training these MLPs with multiview 2D images (without using direct 3D supervision) ~\cite{niemeyer2020differentiable,yariv2020universal,mildenhall2020nerf}.

Most of the existing methods deal with \emph{static scenes}. 
Directly extending the MLPs to encode the additional time dimension does not work well due to 3D shape and motion entanglement. 
The method in \cite{martin2020nerf} extends NeRF for handling crowdsourced photos that contain lighting variations and transient objects. 
Our use of static/dynamic NeRF is similar to that \cite{martin2020nerf}, but we focus on modeling the \emph{dynamic} objects (as opposed to \emph{static} scene in \cite{martin2020nerf}).
The work that most related to ours is \cite{niemeyer2019occupancy}, which learns a continuous motion field over space and time. 
Our work is similar in that we also disentangle the shape/appearance and the motion for dynamic scene elements.
Unlike \cite{niemeyer2019occupancy}, our method models the shape and appearance of a dynamic scene from a casually captured video \emph{without} accessing ground truth 3D information for training.

\topic{Concurrent work on dynamic view synthesis.} Very recently, several methods concurrently to ours have been proposed to extend NeRF for handling dynamic scenes~\cite{xian2020space,li2020neural,tretschk2020non,park2020deformable,pumarola2020d}. 
These methods either disentangle the dynamic scenes into a canonical template and deformation fields for each frame~\cite{tretschk2020non,pumarola2020d,park2020deformable} or directly estimate dynamic (4D spatiotemporal) radiance fields \cite{xian2020space,li2020neural}.
Our work adopts the 4D radiance fields approach due to its capability of modeling large scene dynamics.
In particular, our approach share high-level similarity with \cite{li2020neural} in that we also regularize the dynamic NeRF through scene flow estimation.
Our method differs in several important technical details, including scene flow based 3D temporal consistency loss, sparsity regularization, and the rigidity regularization of the scene flow prediction.
For completeness, we include experimental comparison with one template-based method~\cite{tretschk2020non} and one 4D radiaence field approach~\cite{li2020neural}.









\section{Method}
\label{sec:method}

\subsection{Overview}

\begin{figure*}[t]
\newlength\figwidthstatic
\setlength\figwidthstatic{0.331\linewidth}
\centering%
\parbox[t]{\figwidthstatic}{\centering%
    \includegraphics[width=\figwidthstatic]{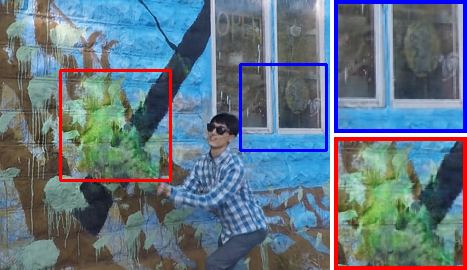}\\%
	(a) NeRF}%
\hfill%
\parbox[t]{\figwidthstatic}{\centering%
  \includegraphics[width=\figwidthstatic]{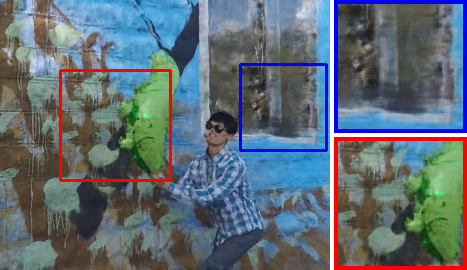}\\%
	(b) NeRF + time}%
\hfill%
\parbox[t]{\figwidthstatic}{\centering%
  \includegraphics[width=\figwidthstatic]{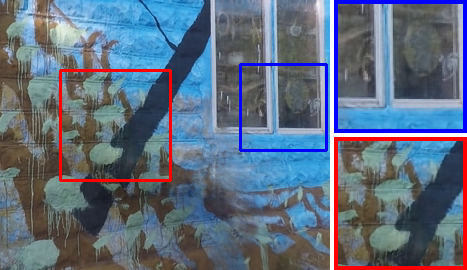}\\%
	(c) Static NeRF (Ours)}%
\vspace{\figcapmargin}
\caption{\textbf{Why static NeRF?}
NeRF~\cite{mildenhall2020nerf} assumes that the scene is entirely static. 
(a) Directly training a NeRF model on a dynamic scene inevitably results in blurry reconstruction (even for the static regions of the scene).
(b) One straightforward extension is to include time as an additional input dimension (NeRF + time). 
However, such a method suffers from ambiguity because the input video can be explained either with time-varying geometry or appearance or both.
The representation reconstructs the input frames well but produces visual artifacts at novel views.
(c) To tackle this issue, we model the static components of the scene using a static NeRF. 
We exclude all the pixels marked as ``dynamic" from training the model. 
This allows us to accurately reconstruct the background's structure and appearance without conflicting the moving objects.
}
\label{fig:static_NeRF}
\end{figure*}
Our method takes as input
(1) monocular video $\left\{\bm{I}_0,\,\bm{I}_1,\,\ldots,\,\bm{I}_{N-1}\right\}$ with $N$ frames, and
(2) a binary mask $\bm{M}$ of the foreground object for each frame.
The mask can be obtained automatically via segmentation or motion segmentation algorithms or semi-automatically via interactive methods such as rotoscoping.
Our goal is to learn a global representation that facilitates free-viewpoint rendering at arbitrary views and input time steps.

Specifically, we build on neural radiance fields (NeRFs)~\cite{mildenhall2020nerf} as our base representation. 
NeRF models the scene implicitly with a continuous and differentiable function (i.e., an MLP) that regresses an input 3D position $\x = (x, y, z)$ and the normalized viewing direction $\d = (d_x, d_y, d_z)$ to the corresponding volume density $\sigma$ and color $\c = (r,g,b)$.
Such representations have demonstrated high-quality view synthesis results when trained with multiple images of a scene.
However, NeRF assumes that the scene is \emph{static} (with constant density and radiance). 
This assumption does not hold for casually captured videos of dynamic scenes.

One straightforward extension of the NeRF model would be to include \emph{time} as an additional dimension as input, e.g., using 4D position $(x, y, z, t)$ input where $t$ denotes the index of the frame.
While this model theoretically can represent the time-varying structure and appearance of a dynamic scene, the model training is highly ill-posed, given that we only have one single 2D image observation at each time step.
There exist infinitely many possible solutions that match the input video exactly.
Empirically, we find that directly training the ``NeRF + time'' model leads to low visual quality.

The key contribution of our paper lies in resolving this ambiguity for modeling the time-varying radiance fields.
To this end, we propose to use different models to represent \emph{static} or \emph{dynamic} scene components using the user-provided dynamic masks. 

For \emph{static components} of the scene, we apply the original NeRF model~\cite{mildenhall2020nerf}, but exclude all ``dynamic" pixels from training the model. 
This allows us to reconstruct the background's structure and appearance without conflicting reconstruction losses from moving objects.
We refer to this model as ``Static NeRF" (\figref{static_NeRF}).

\begin{figure}[t]
\newlength\figwidthdynamic
\setlength\figwidthdynamic{0.482\linewidth}
\newlength\figwidthcaption
\setlength\figwidthcaption{0.02\linewidth}
\centering%
\mpage{0.01}{\rotatebox[origin=c]{90}{ Novel view \quad Reconstruction}}
\parbox[t]{\figwidthdynamic}{\centering%
 \includegraphics[trim=100 40 0 0, clip=true, width=\figwidthdynamic]{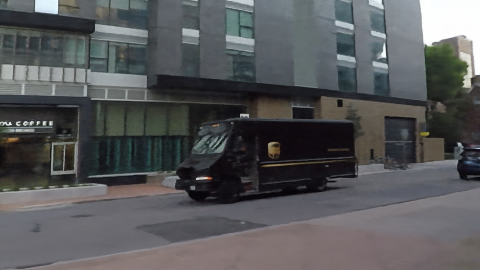}\\%
 \includegraphics[trim=0 0 0 0, clip=true, width=\figwidthdynamic]{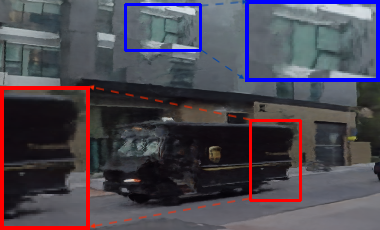}\\%
	(a) NeRF + time}%
\hfill%
\parbox[t]{\figwidthdynamic}{\centering%
  \includegraphics[trim=100 40 0 0, clip=true, width=\figwidthdynamic]{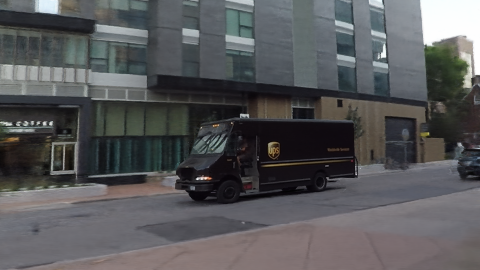}\\%
  \includegraphics[trim=0 0 0 0, clip=true, width=\figwidthdynamic]{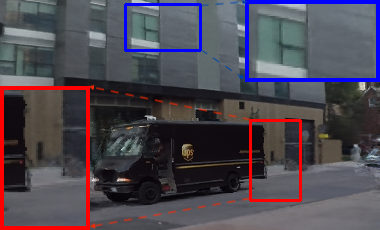}\\%
	(b) Ours}%
\vspace{\figcapmargin}
\caption{\textbf{Why dynamic NeRF?}
(\emph{Top}) Since the training objective is to minimize the image reconstruction loss on the input video frames, NeRF + time explains the input frames very well.
(\emph{Bottom}) However, there are infinitely many solutions that can correctly render the given input video, yet only one of them is physically correct for generating photorealistic novel views.
NeRF + time tries to disentangle view from time using time as additional input. 
However, the problem becomes under-constrained and leads to artifacts in both static and dynamic regions.
Our dynamic NeRF produces plausible view synthesis results for moving objects.
}
\label{fig:dynamic_NeRF}
\end{figure}

For \emph{dynamic components} of the scene (e.g., moving objects), we train an MLP that takes a 3D position and time $(x, y, z, t)$ as input to model the volume density and color of the dynamic objects at each time instance.
To leverage the multi-view geometry, we use the same MLP to predict the additional three-dimensional scene flow from time $t$ to the previous and next time instance.
Using the predicted forward and backward scene flow, we create a \emph{warped} radiance field (similar to the backward warping 2D optical flow) by resampling the radiance fields implicitly modeled at time $t+1$ and $t-1$.
For each 3D position, we then have up to three multi-view observations to train our model.
We refer to this model as ``Dynamic NeRF" (\figref{dynamic_NeRF}).
Additionally, our Dynamic NeRF predicts a blending weight and learns how to blend the results from both the static NeRF and dynamic NeRF in an unsupervised manner.
In the following, we discuss the detailed formulation of the proposed static and dynamic NeRF models and the training losses for optimizing the weights for the implicit functions.



\subsection{Static NeRF}
\label{sec:static_NeRF}
\topic{Formulation.}
Our static NeRF follows closely the formulation in \cite{mildenhall2020nerf}
and is represented by a fully-connected neural network\footnote{We refer the readers to \cite{mildenhall2020nerf} for implementation details.}.
Consider a ray from the camera center $\mathbf{o}$ through a given pixel on the image plane as $\r(\interval_k) = \mathbf{o} + \interval_k \mathbf{d}$, where $\mathbf{d}$ is the normalized viewing direction, our static NeRF maps a 3D position $\r(\interval_k)$ and viewing direction $\d$ to volume density $\sigma^s$ and color $\cs$:
\vspace{\eqnmargin}
\begin{equation}
\left( \sigma^s, \cs \right) = \text{MLP}_{\!\theta} \left( \r(\interval_k) \right), 
\label{eq:staticNeRF}
\end{equation}
where $\text{MLP}_{\!\theta}$ stands for two cascaded MLP, detailed in \figref{overview}.
We can compute the color of the pixel (corresponding the ray $\r(\interval_k)$) using numerical quadrature for approximating the volume rendering interval~\cite{drebin1988volume}:
\vspace{\eqnmargin}
\begin{align}
\mathbf{C}^s(\r) &= \sum_{k=1}^{K} T^s\!(\interval_k) \, \alpha^s\!( \sigma^s\!(\interval_k)\,\delta_k )\, \mathbf{c}^s\!(\interval_k), \\
T^s\!(\interval_k) &= \exp\!\left( - \sum_{k'=1}^{k-1} \sigma^s\!(\interval_k)\,\delta_k \right),
\label{eq:static_rendering}
\end{align}
where $\alpha(x) = 1 - \exp\!\left(- x \right)$ and $\delta_k = \interval_{k+1} - \interval_k$ is the distance between two quadrature points. 
The $K$ quadrature points $\{\interval_k\}_{k=1}^K$ are drawn uniformly between $\interval_n$ and $\interval_{\!f}$ \cite{mildenhall2020nerf}.
$T^s\!(\interval_k)$ indicates the accumulated transmittance from $\interval_n$ to $\interval_k$.

\topic{Static rendering photometric loss.}
To train the weights $\theta_s$ of the static NeRF model, we first construct the camera rays using all the pixels for all the video frames (using the associated intrinsic and extrinsic camera poses for each frame).
Here we denote $\r_{ij}$ as the rays passing through the pixel $j$ on image $i$ with $\r_{ij}(\interval) = \mathbf{o}_i + (\interval) \mathbf{d}_{ij}$.
We can then optimize $\theta_s$ by minimizing the \emph{static rendering photometric loss} for all the color pixels $\mathbf{C}(\r_{ij})$ in frame $i\in \{0,\,\ldots,\,N\!-\!1 \}$ in the static regions (where $\mathbf{M}(\r_{ij}) = 0$):
\begin{align}
\mathcal{L}_{static} = \sum_{ij} \norm{ ( \mathbf{C}^s(\r_{ij}) - \mathbf{C}^\textit{gt}(\r_{ij})) \cdot ( 1 - \mathbf{M}(\r_{ij})) }^2_2
\label{eq:static_loss}
\end{align}

\subsection{Dynamic NeRF}
\label{sec:dynamic_NeRF}
In this section, we introduce our core contribution to modeling time-varying radiance fields using \emph{dynamic NeRF}.
The challenge lies in that we only have \emph{one single} 2D image observation at each time instance $t$. 
So the training lacks multi-view constraints.
To resolve this training difficulty, we predict the forward and backward scene flow and use them to create a \emph{warped} radiance field by resampling the radiance fields implicitly modeled at time $t+1$ and $t-1$.
For each 3D position at time $t$, we then have up to three 2D image observations.
This multi-view constraint effectively constrains the dynamic NeRF to produce temporally consistent radiance fields. 

\topic{Formulation.}
Our dynamic NeRF takes a 4D-tuple $(\r(\interval_k),\,t)$ as input and predict 3D scene flow vectors $\s_\textit{fw}$, $\s_\textit{bw}$, volume density $\sigma^d$, color $\cd$ and blending weight $b$:
\vspace{\eqnmargin}
\begin{equation}
\left( \s_\textit{fw}, \s_\textit{bw}, \sigma^d_t, \cd_t, b \right) = \text{MLP}_{\theta_d} \left( \r(\interval_k), t \right)
\label{eq:dynamicNeRF}
\end{equation}

Using the predicted scene flow $\s_\textit{fw}$ and $\s_\textit{bw}$, we obtain the scene flow neighbors $\r(\interval_k) + \s_\textit{fw}$ and $\r(\interval_k) + \s_\textit{bw}$.
We also use the predicted scene flow to \emph{warp} the radiance fields from the neighboring time instance to the current time.
For every 3D position at time $t$, we obtain the occupancy $\sigma^d$ and color $\cd$ through querying the same MLP model at $\r(\interval_k) + \s$:
\vspace{\eqnmargin}
\begin{align}
\left( \sigma^d_{\!t+1},\,\cd_{t+1} \right) &= \text{MLP}_{\theta_d}\!\left( \r(\interval_k) + \s_\textit{fw},\,t + 1 \right) \\
\left( \sigma^d_{\!t-1},\,\cd_{t-1} \right) &= \text{MLP}_{\theta_d}\!\left( \r(\interval_k) + \s_\textit{bw},\,t - 1 \right)
\label{eq:dynamicNeRF_warp}
\end{align}

For computing the color of a dynamic pixel at time $t'$, we use the following approximation of volume rendering integral: 
\vspace{\eqnmargin}
\begin{align}
\mathbf{C}^d_{t'}\!(\r) &= \sum_{k=1}^{K} T^d_{t'}(\interval_k)\,\alpha^d\!( \sigma^d_{t'}(\interval_k)\,\delta_k ) \, \mathbf{c}^d_{t'}(\interval_k) 
\label{eq:dynamic_rendering}
\end{align}

\topic{Dynamic rendering photometric loss.}
Similar to the static rendering loss, we train the dynamic NeRF model by minimizing the reconstruction loss:
\vspace{\eqnmargin}
\begin{align}
\mathcal{L}_\textit{dyn} = \sum_{t' \in \{ t,\,t-1,\,t+1\}}\sum_{ij} \norm{ ( \mathbf{C}^d_{t'}(\r_{ij}) - \mathbf{C}^\textit{gt}(\r_{ij}))}^2_2
\label{eq:dynamic_loss}
\end{align}

\subsection{Regularization Losses for Dynamic NeRF}
While leveraging the multi-view constraint in the dynamic NeRF model reduces the amount of ambiguity, the model training remains ill-posed without proper regularization. 
To this end, we design several regularization losses to constrain the Dynamic NeRF.

\begin{figure}[t]
\newlength\figwidthFlow
\setlength\figwidthFlow{0.328\linewidth}
\parbox[t]{\figwidthFlow}{\centering%
  \fbox{\includegraphics[trim=278 140 10 10, clip=true, width=\figwidthFlow]{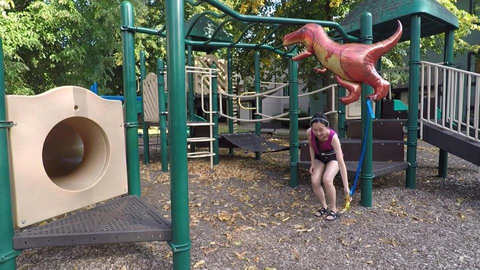}}\\
  \small (a) Input}%
\hfill
\parbox[t]{\figwidthFlow}{\centering%
  \fbox{\includegraphics[trim=278 140 10 10, clip=true, width=\figwidthFlow]{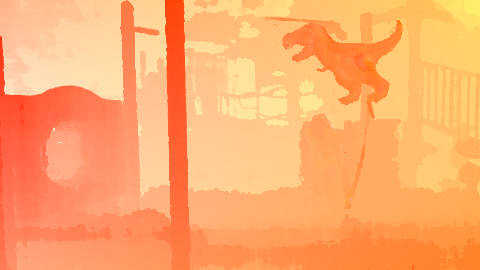}}\\
  \small (b) Induced flow}%
\hfill
\parbox[t]{\figwidthFlow}{\centering%
  \fbox{\includegraphics[trim=278 140 10 10, clip=true, width=\figwidthFlow]{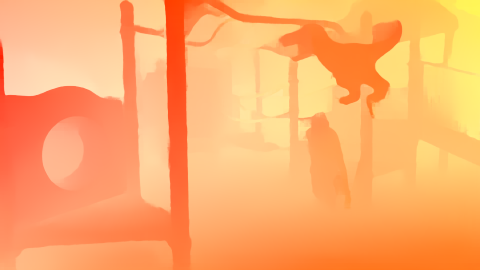}}\\
  \small (c) Estimated flow}%
\vspace{\figcapmargin}
\caption{\textbf{Scene flow induced optical flow.}
We supervise the predicted scene flow by minimizing the endpoint error between the estimated optical flow~\cite{Teed-RAFT-ECCV} and our scene flow induced optical flow.
Since we jointly train our model with \emph{both} photometric loss and motion matching loss, our learned volume density helps render a more accurate flow than the estimated flow (e.g., the complex structures of the fence on the right).
}
\label{fig:flow}
\end{figure}

\topic{Motion matching loss.}
As we do not have direct 3D supervision for the predicted scene flow from the motion MLP model, we use 2D optical flow (estimated from input image pairs using~\cite{Teed-RAFT-ECCV}) as \emph{indirect} supervision.
For each 3D point at time $t$, we first use the estimated scene flow to obtain the corresponding 3D point in the reference frame. 
We then project this 3D point onto the reference camera so we can compute the \emph{scene flow induced optical flow} and enforce it to match the estimated optical flow (\figref{flow}). 
Since we jointly train our model with both photometric loss and motion matching loss, the learned volume density helps render a more accurate flow than the estimated flow. Thus, we do not suffer from inaccurate optical flow supervision.

\begin{figure}[t]
\begin{overpic}[width=1\linewidth]{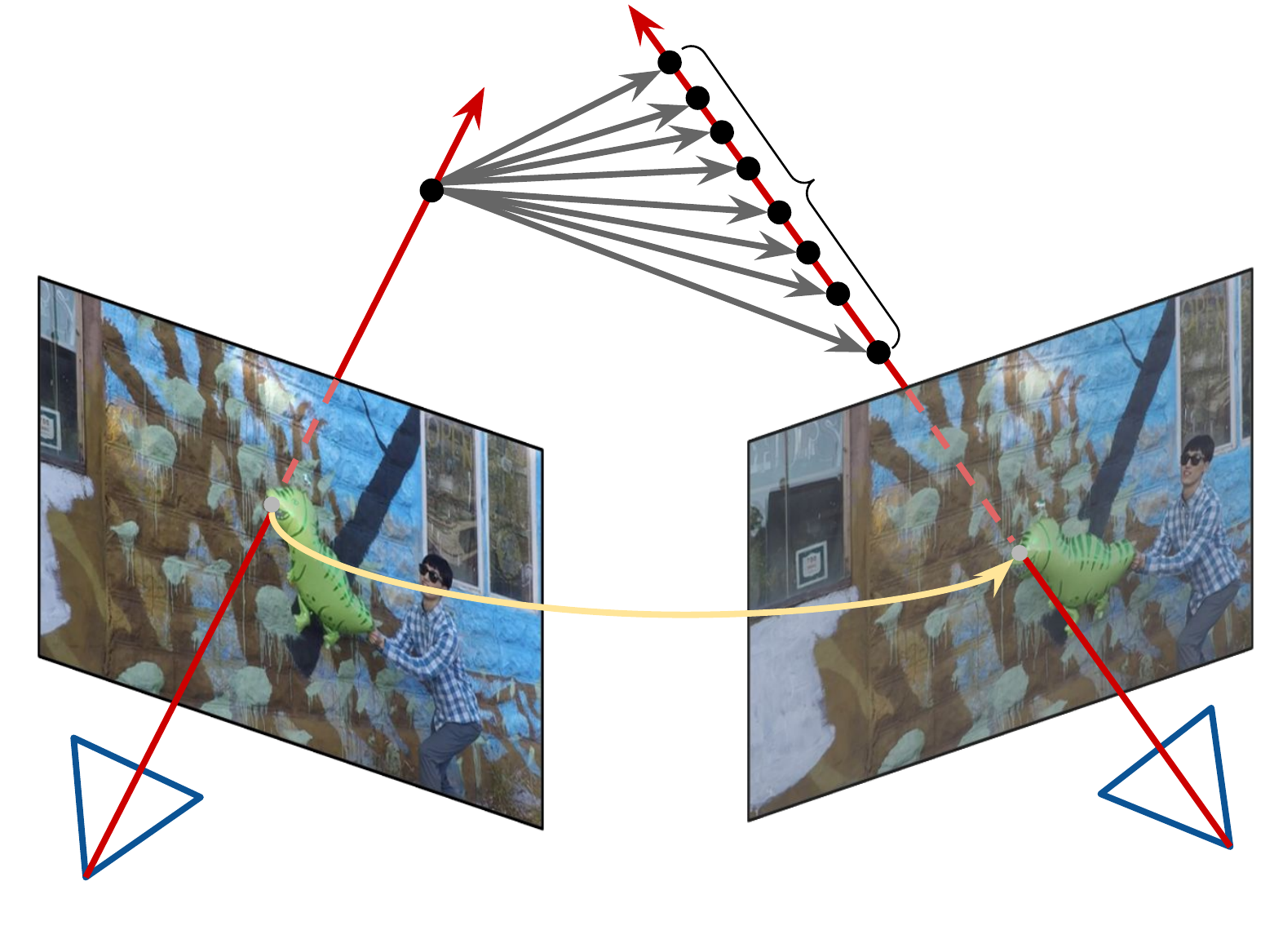}
\put (16, 59) {$(\r(\interval_k), t)$}
\put (65, 59) {$(\r(\interval_k) + \sfw, t + 1)$}
\put (16, 5) {$t$}
\put (80, 5) {$t + 1$}
\put (44.5, 20) {Optical}
\put (46.2, 15.0) {flow}
\end{overpic}
\vspace{\figcapmargin}
\caption{
\textbf{Ambiguity of optical flow supervision.}
Matching the scene flow induced optical flow with the estimated 2D optical flow does not fully resolve the ambiguity. 
There exists a 1D family of scene flow predictions that produce the same 2D optical flow.
}
\label{fig:motivation_sceneflow_reg}
\end{figure}
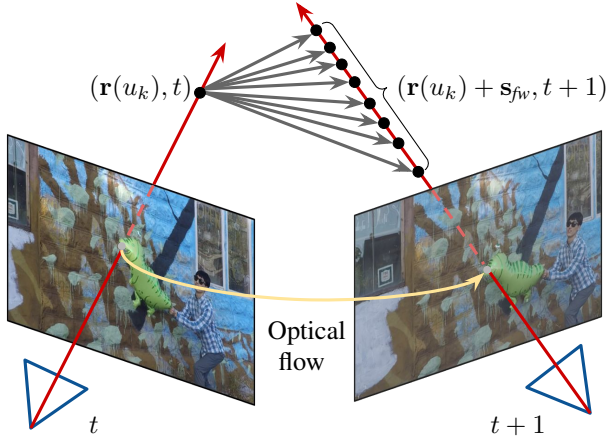

\topic{Motion regularization.}
Unfortunately, matching the rendered scene flow with 2D optical flow does not fully resolve all ambiguity, as a 1D family of scene flow vectors produces the same 2D optical flow (\figref{motivation_sceneflow_reg}).
We regularize the scene flow to be \emph{slow} and \emph{temporally smooth}:
\vspace{\eqnmargin}
\begin{align}
\mathcal{L}_\textit{slow} &= \sum_{ij} \norm{ \sfw(\rij) }_1 +  \norm{ \sbw(\rij) }_1 \\
\vspace{\eqnmargin}
\mathcal{L}_\textit{smooth} &= \sum_{ij} \norm{  \sfw(\rij) +  \sbw(\rij) }_2^2
\label{eq:slow_loss}
\end{align}
\vspace{\eqnmargin}

We further regularize the scene flow to be \emph{spatially smooth} by minimizing the difference between neighboring 3D points' scene flow. 
To regularize the consistency of the scene flow, we have the scene flow cycle consistency regularization:
\vspace{\eqnmargin}
\begin{align}
\mathcal{L}_\textit{cyc} = \sum &\norm{ \sfw(\r, t) + \sbw(\r + \sfw(\r, t),\,t+1) }_2^2\\
+ &\norm{ \sbw(\r, t) + \sfw(\r + \sbw(\r, t),\,t-1) }_2^2
\label{eq:consistency_loss}
\end{align}

\topic{Sparsity regularization.}
We render the color using principles from classical volume rendering.
One can see through a particle if it is partially transparent.
However, one can not see through the scene flow because the scene flow is not an intrinsic property (unlike color).
Thus, we minimize the entropy of the rendering weights $T^d \alpha^d$ along each ray so that few samples dominate the rendering. 

\topic{Depth order loss.} 
For a moving object, we can either interpret it as an object close to the camera moving slowly or an object far away moving fast.
To resolve the ambiguity, we leverage the state-of-the-art single-image depth estimation~\cite{lasinger2019towards} to estimate the input depth. 
As the depth estimates are up to shift and scale, we cannot directly use them to supervise our model. 
Instead, we use the robust loss as in~\cite{lasinger2019towards} to constrain our dynamic NeRF.
Since the supervision is still up to shift and scale, we further constrain our dynamic NeRF with our static NeRF.
Thanks to the multi-view constraint, our static NeRF estimates accurate depth for the static region. We additionally minimize the L2 difference between $\mathbf{D}^\textit{s}$ and $\mathbf{D}^\textit{d}$ for all the pixels in the static regions (where $\mathbf{M}(\r_{ij}) = 0$):
\vspace{\eqnmargin}
\begin{align*}
  \mathcal{L}_\textit{depth} = \sum_{ij} &\norm{ \overline{\mathbf{D}^d}(\r_{ij})-  \overline{\mathbf{D}^\textit{gt}}(\r_{ij})  }^2_2 + \\
  &\norm{ ( \mathbf{D}^d(\r_{ij}) - \mathbf{D}^\textit{s}(\r_{ij})) \cdot ( 1 - \mathbf{M}(\r_{ij})) }^2_2,
\end{align*}
where $\overline{\mathbf{D}}$ stands for the normalized depth.


\topic{3D temporal consistency loss.}
If an object remains unmoved for a while, the network can not learn the correct volume density and color of the \emph{occluded background} at the current time because those 3D positions are omitted during volume rendering.
When rendering a novel view, the model may generate holes for the occluded region. 
To address this issue, we propose the 3D temporal consistency loss \emph{before} rendering. 
Specifically, we enforce the volume density and color of each 3D position to match its scene flow neighbors'.
The correct volume density and color will then be \emph{propagated} across time steps.

\topic{Rigidity regularization of the scene flow.}
Our model prefers to explain a 3D position by the static NeRF if this position has no motion.
For static position, we want the blending weight $b$ to be closed to $1$.
For a non-rigid position, the blending weight $b$ should be $0$.
This learned blending weight can further constrain the rigidity of the predicted scene flow by taking the product of the predicted scene flow and $(1 - b)$.
If a 3D position has no motion, the scene flow is forced to be zero.

\begin{figure}[t]
\newlength\figwidthDS
\setlength\figwidthDS{0.312\linewidth}
\centering%
\mpage{0.02}{\rotatebox[origin=c]{90}{Depth \quad \quad Color \quad}} \hfill%
%
\hfill%
\parbox[t]{\figwidthDS}{\centering%
  \includegraphics[trim=0 0 0 0, clip=true, width=\figwidthDS]{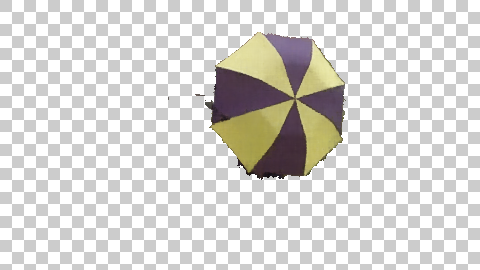}\\%
  \includegraphics[trim=0 0 0 0, clip=true, width=\figwidthDS]{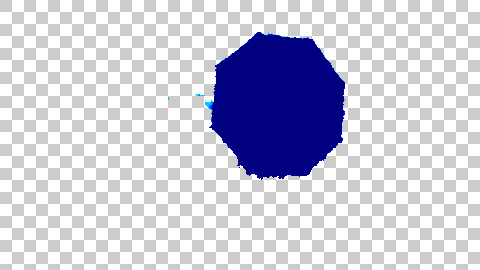}\\%
  \small  (a) Dynamic NeRF}%
\hfill%
\parbox[t]{\figwidthDS}{\centering%
 \includegraphics[trim=0 0 0 0, clip=true, width=\figwidthDS]{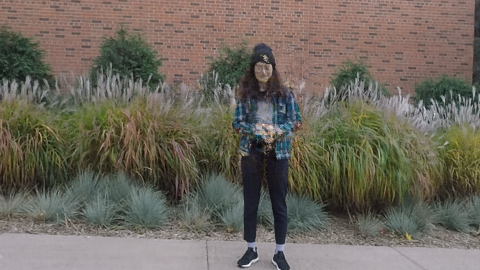}\\%
 \includegraphics[trim=0 0 0 0, clip=true, width=\figwidthDS]{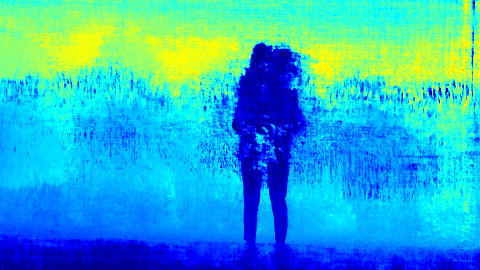}\\%
 \small (b) Static NeRF}%
\hfill%
\parbox[t]{\figwidthDS}{\centering%
  \includegraphics[trim=0 0 0 0, clip=true, width=\figwidthDS]{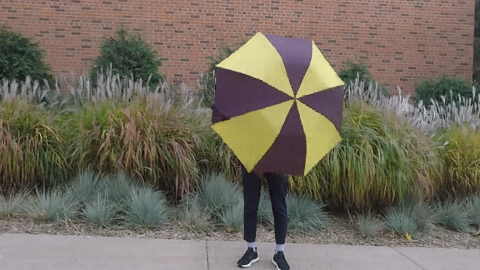}\\%
  \includegraphics[trim=0 0 0 0, clip=true, width=\figwidthDS]{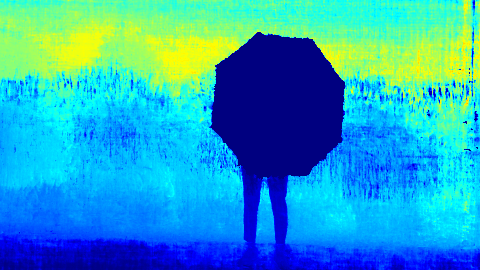}\\%
  \small (c) Full model}%
\vspace{\figcapmargin}
\caption{\textbf{Full model rendering.}
We compose the (a) dynamic and (b) static NeRF model into (c) our full model and render full frames at novel viewpoints and time steps.
}
\label{fig:DSFull}
\end{figure}
\begin{table*}[t]
\caption{
\textbf{Novel view synthesis results.}
We report the average PSNR and LPIPS results with comparisons to existing methods on Dynamic Scene dataset~\cite{Yoon-2020-CVPR}.
The best performance is in \first{bold} and the second best is \second{underscored}.
}
\label{tab:results}
\centering
\resizebox{1\linewidth}{!} 
{
\begin{tabular}{l ccccccc | c}
\toprule
PSNR $\uparrow$ / LPIPS $\downarrow$ & Jumping & Skating & Truck & Umbrella & Balloon1 & Balloon2 & Playground & Average \\
\midrule
NeRF & 
20.58 / 0.305 &
23.05 / 0.316 &
22.61 / 0.225 &
21.08 / 0.441 &
19.07 / 0.214 &
24.08 / 0.098 &
20.86 / \second{0.164} &
21.62 / 0.252 \\
NeRF + time & 
16.72 / 0.489 &
19.23 / 0.542 &
17.17 / 0.403 &
17.17 / 0.752 &
17.33 / 0.304 &
19.67 / 0.236 &
13.80 / 0.444 &
17.30 / 0.453 \\
Yoon~\etal~\cite{Yoon-2020-CVPR} & 
20.16 / \second{0.148} & 
21.75 / \second{0.135} & 
23.93 / \first{0.109} & 
20.35 / \second{0.179} & 
18.76 / \second{0.178} & 
19.89 / \second{0.138} & 
15.09 / 0.183 & 
19.99 / \second{0.153} \\
Tretschk~\etal~\cite{Tretschk-NR} & 
19.38 / 0.295 &
23.29 / 0.234 &
19.02 / 0.453 &
19.26 / 0.427 &
16.98 / 0.353 &
22.23 / 0.212 &
14.24 / 0.336 &
19.20 / 0.330 \\
Li~\etal~\cite{Li-NSFF} & 
\second{24.12} / 0.156 &
\first{28.91} / \second{0.135} &
\first{25.94} / 0.171 &
\second{22.58} / 0.302 &
\second{21.40} / 0.225 &
\second{24.09} / 0.228 &
\second{20.91} / 0.220 &
\second{23.99} / 0.205 \\
Ours & 
\first{24.23} / \first{0.144} & 
\second{28.90} / \first{0.124} & 
\second{25.78} / \second{0.134} & 
\first{23.15} / \first{0.146} & 
\first{21.47} / \first{0.125} & 
\first{25.97} / \first{0.059} & 
\first{23.65} / \first{0.093} & 
\first{24.74} / \first{0.118} \\
\bottomrule
\end{tabular}
}
\end{table*}
\begin{figure*}[t]
\newlength\figwidthNovel
\setlength\figwidthNovel{0.163\linewidth}
\centering%
%
%
\parbox[t]{\figwidthNovel}{\centering%
  \includegraphics[trim=120 50 0 20, clip=true, width=\figwidthNovel]{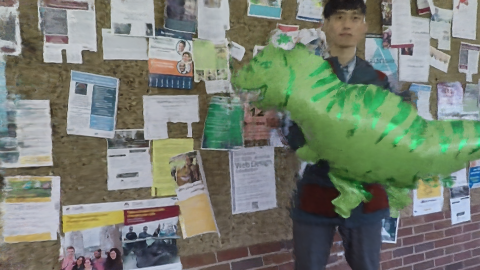}\\%
  \includegraphics[trim=250 70 0 70, clip=true, width=\figwidthNovel]{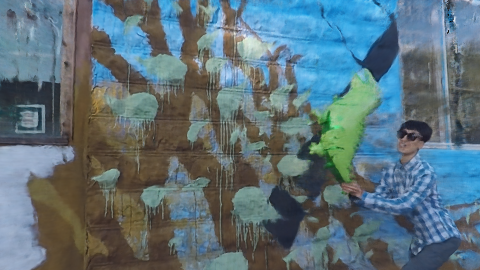}\\%
  \includegraphics[trim=70 40 50 30, clip=true, width=\figwidthNovel]{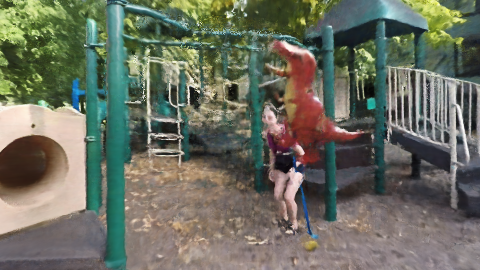}\\%
  \includegraphics[trim=180 70 0 20, clip=true, width=\figwidthNovel]{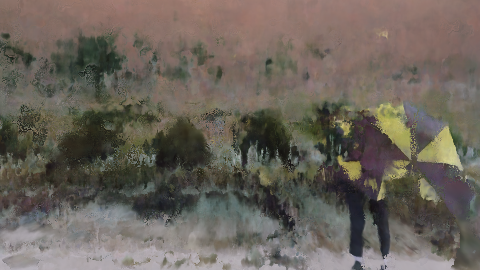}\\%
	NeRF + time}%
\hfill%
\parbox[t]{\figwidthNovel}{\centering%
  \includegraphics[trim=120 50 0 20, clip=true, width=\figwidthNovel]{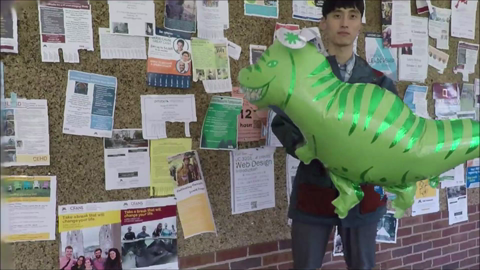}\\%
  \includegraphics[trim=250 70 0 70, clip=true, width=\figwidthNovel]{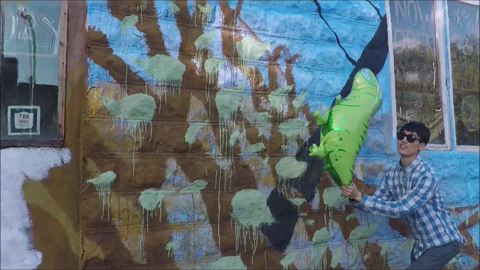}\\%
  \includegraphics[trim=70 40 50 30, clip=true, width=\figwidthNovel]{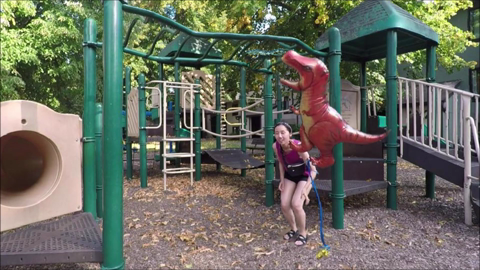}\\%
  \includegraphics[trim=180 70 0 20, clip=true, width=\figwidthNovel]{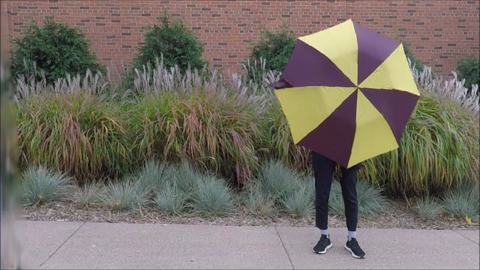}\\%
  Yoon~\etal~\cite{Yoon-2020-CVPR}}%
\hfill%
\parbox[t]{\figwidthNovel}{\centering%
  \includegraphics[trim=120 50 0 20, clip=true, width=\figwidthNovel]{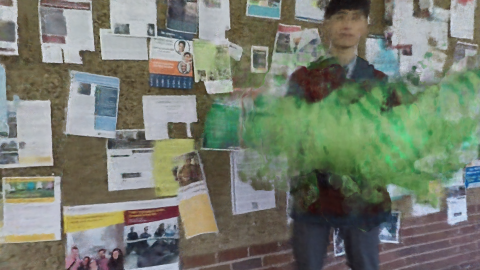}\\%
  \includegraphics[trim=250 70 0 70, clip=true, width=\figwidthNovel]{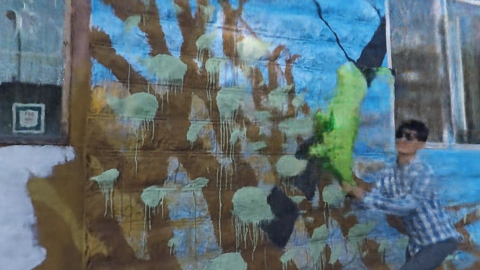}\\%
  \includegraphics[trim=70 40 50 30, clip=true, width=\figwidthNovel]{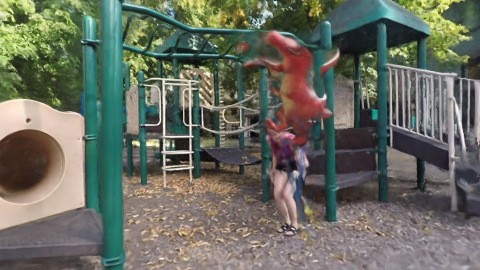}\\%
  \includegraphics[trim=180 70 0 20, clip=true, width=\figwidthNovel]{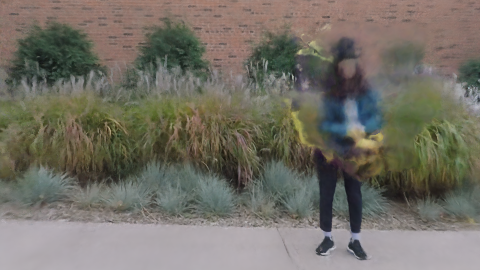}\\%
  Tretschk~\etal~\cite{Tretschk-NR}}%
\hfill%
\parbox[t]{\figwidthNovel}{\centering%
  \includegraphics[trim=120 50 0 20, clip=true, width=\figwidthNovel]{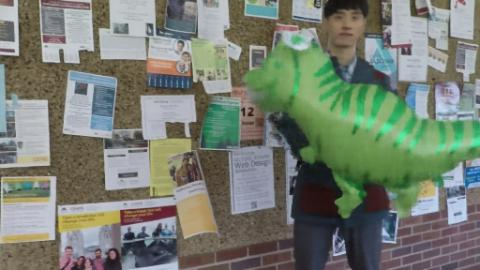}\\%
  \includegraphics[trim=250 70 0 70, clip=true, width=\figwidthNovel]{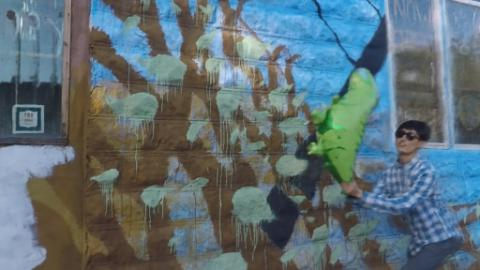}\\%
  \includegraphics[trim=70 40 50 30, clip=true, width=\figwidthNovel]{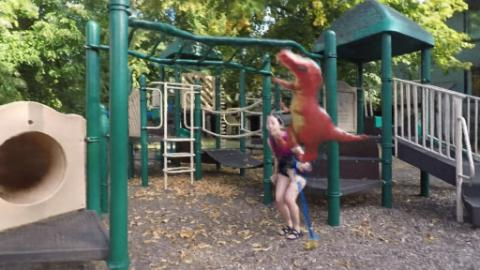}\\%
  \includegraphics[trim=180 70 0 20, clip=true, width=\figwidthNovel]{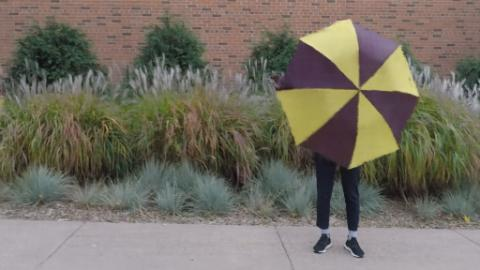}\\%
  Li~\etal~\cite{Li-NSFF}}%
\hfill%
\parbox[t]{\figwidthNovel}{\centering%
  \includegraphics[trim=120 50 0 20, clip=true, width=\figwidthNovel]{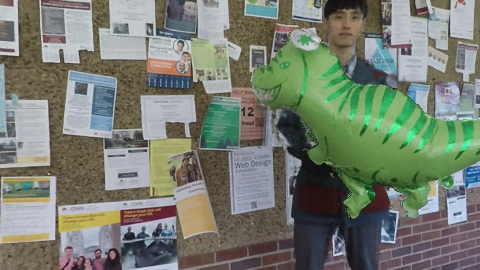}\\%
  \includegraphics[trim=250 70 0 70, clip=true, width=\figwidthNovel]{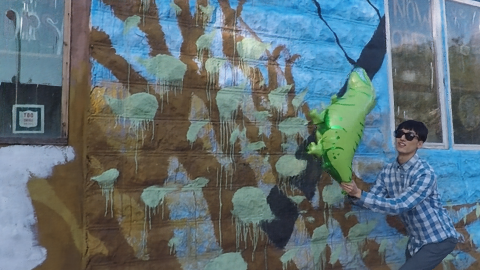}\\%
  \includegraphics[trim=70 40 50 30, clip=true, width=\figwidthNovel]{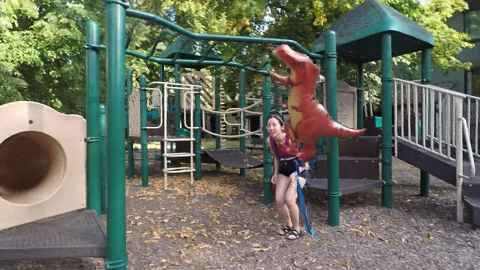}\\%
  \includegraphics[trim=180 70 0 20, clip=true, width=\figwidthNovel]{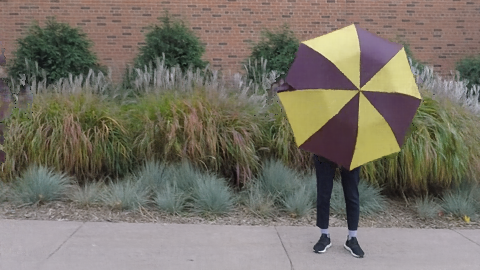}\\%
  Ours}%
\hfill%
\parbox[t]{\figwidthNovel}{\centering%
  \includegraphics[trim=120 50 0 20, clip=true, width=\figwidthNovel]{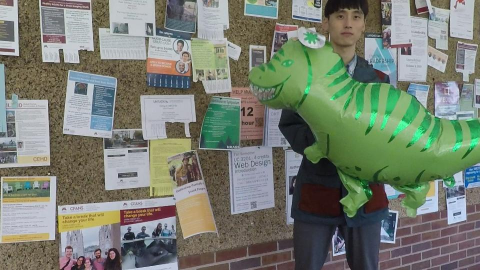}\\%
  \includegraphics[trim=250 70 0 70, clip=true, width=\figwidthNovel]{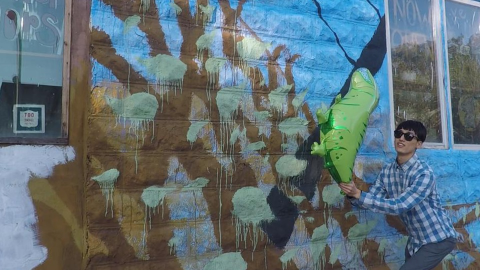}\\%
  \includegraphics[trim=70 40 50 30, clip=true, width=\figwidthNovel]{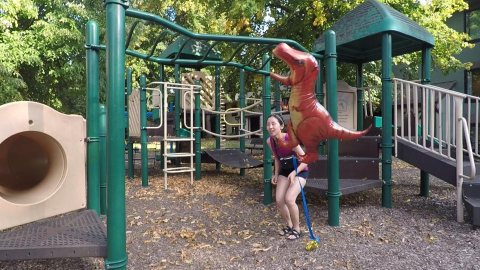}\\%
  \includegraphics[trim=180 70 0 20, clip=true, width=\figwidthNovel]{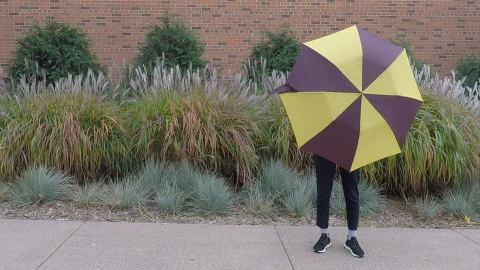}\\%
  Ground truth}%
\vspace{\figcapmargin}
\caption{\textbf{Novel view synthesis.} 
Our model enables the free-viewpoint synthesis of a dynamic scene. 
Compared with Yoon~\etal~\cite{Yoon-2020-CVPR}, our results appear slightly blurry (because we reconstruct the entire frame as opposed to warp and blend input images), but align with the ground truth image better and create smoother view-interpolation results. 
When compared to other NeRF-based methods, our results are sharper and closer to the ground truth.
Please refer to the supplementary material for video results.
}
\label{fig:novel_views}
\end{figure*}
\subsection{Combined model}
With both the static and dynamic NeRF model, we can easily compose them into a complete model using the predicted blending weight $b$ and render full color frames at novel views and time:
\vspace{\eqnmargin}
\begin{align}
\mathbf{C}^\textit{full}(\mathbf{r}) &= \sum_{k=1}^{K} T^\textit{full}\!\Big( \alpha^d ( \sigma^d\delta_k ) (1 - b) \mathbf{c}^d + \alpha^s ( \sigma^s\delta_k ) b \mathbf{c}^s \Big)
\label{eq:full_rendering}
\end{align}
We predict the blending weight $b_d$ using the dynamic NeRF to enforce the time-dependency.
Using the blending weight, we can also render a dynamic component only frame where the static region is transparent (\figref{DSFull}).

\topic{Full rendering photometric loss.}
We train the two NeRF models jointly by applying a reconstruction loss on the composite results:
\vspace{\eqnmargin}
\begin{align}
\mathcal{L}_\textit{full} = \sum_{ij} \norm{ \mathbf{C}^\textit{full}(\mathbf{r}_{ij}) - \mathbf{C}^\textit{gt}(\mathbf{r}_{ij}) }^2_2
\label{eq:full_loss}
\end{align}

\section{Experimental Results}
\label{sec:results}

\subsection{Experimental setup}
\topic{Dataset.} 
We evaluate our method on the Dynamic Scene Dataset \cite{Yoon-2020-CVPR}, which contains 9 video sequences. 
The sequences are captured with 12 cameras using a static camera rig.
All cameras simultaneously capture images at 12 different time steps $\left\{t_0,\,t_1,\,\ldots,\,t_{11}\right\}$.
The input twelve-frames monocular video $\left\{\bm{I}_0,\,\bm{I}_1 ,\,\ldots,\,\bm{I}_{11}\right\}$ is obtained by sampling the image taken by the $i$-th camera at time $t_i$. 
Please note that a different camera is used for each frame of the video to simulate camera motion.
The frame $\bm{I}_i$ contains a background that does not change in time, and a time-varying dynamic object.
Like NeRF~\cite{mildenhall2020nerf}, we use COLMAP to estimate the camera poses and the near and far bounds of the scene.
We assume all the cameras share the same intrinsic parameter.
We exclude the DynamicFace sequence because COLMAP fails to estimate camera poses.
We resize all the sequences to $480 \times 270$ resolution.

\vspace{-1mm}
\subsection{Evaluation}
\topic{Quantitative evaluation.}
To quantitatively evaluate the synthesized novel views, we fix the view to the first camera and change time.
We show the PSNR and LPIPS~\cite{zhang2018unreasonable} between the synthesized views and the corresponding ground truth views in \tabref{results}.
We obtain the results of Li~\etal~\cite{Li-NSFF} and Tretschk~\etal~\cite{Tretschk-NR} using the official implementation with default parameters.
Note that the method from Tretschk~\etal~\cite{Tretschk-NR} needs per-sequence hyper-parameter tuning.
The visual quality might be improved with careful hyper-parameter tuning.
Our method compares favorably against the state-of-the-art algorithms.

\begin{figure}[t]
\newlength\figwidthLi
\setlength\figwidthLi{0.49\linewidth}
\centering%
\includegraphics[width=1\linewidth]{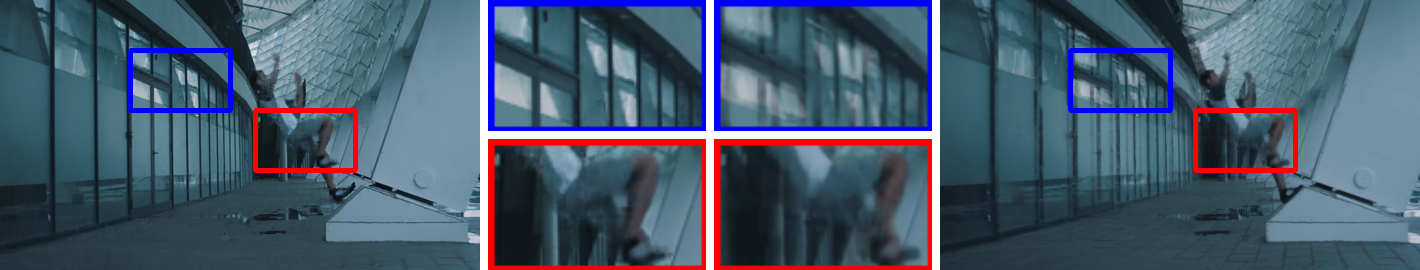}\hfill
\includegraphics[width=1\linewidth]{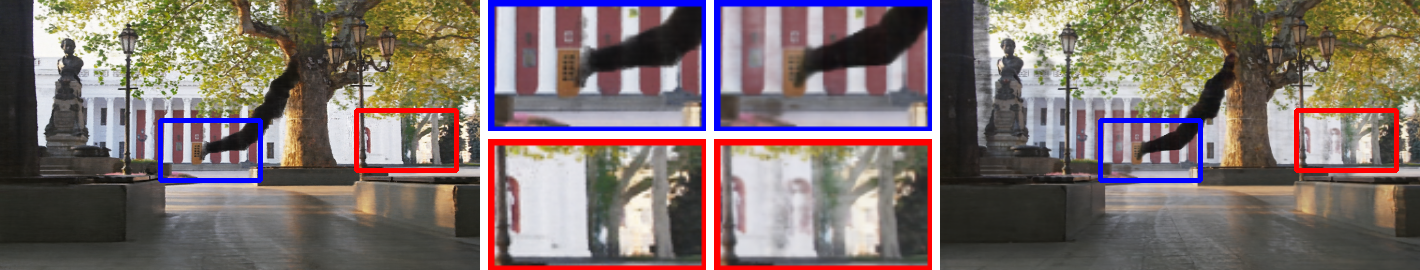}\hfill

\vspace{-4mm}
$\underbracket[1pt][2.0mm]{\hspace{\figwidthLi}}_%
    {\substack{\vspace{-2.0mm}\\\colorbox{white}
    {Ours}}}$\vspace{1mm} \hfill
$\underbracket[1pt][2.0mm]{\hspace{\figwidthLi}}_%
    {\substack{\vspace{-2.0mm}\\\colorbox{white}
    {Li~\etal~\cite{Li-NSFF}}}}$\vspace{1mm}
\vspace{\figcapmargin}
\caption{\textbf{Comparison with~\cite{Li-NSFF}.}
We show that our proposed regularizations are the keys to better visual results.
}
\label{fig:compare_Li}
\end{figure}

\topic{Qualitative evaluation.}
We show the sample view synthesis results in \figref{novel_views}.
With the learned neural implicit representation of the scene, our method can synthesize \emph{novel views} that are never seen during training.
Please refer to the supplementary video results for the novel view synthesis, and the extensive qualitative comparison to the methods listed in \tabref{results}.

\figref{compare_Li} shows the comparison with Li~\etal~\cite{Li-NSFF} on large motion sequences taken in the wild. 
Unlike~\cite{Li-NSFF} which predicts the blending weight using a static NeRF, we learn a time-varying blending weight.
This weight helps better distinguish the static region and yields a clean background.
Our rigidity regularization encourages the scene flow to be zero for the rigid region. 
As a result, the multi-view constraints enforce the background to be static. Without this regularization, the background becomes time-variant and leads to floating artifacts in~\cite{Li-NSFF}.

\begin{table}[t]
    \caption{
    \textbf{Ablation study on different losses.}
    We report PSNR, SSIM and LPIPS on the Playground sequence. 
    }
    \label{tab:ablation}
    \centering
    {
    \begin{tabular}{l | ccc}
    \toprule
    & PSNR $\uparrow$ & SSIM $\uparrow$ & LPIPS $\downarrow$ \\
    \midrule
    %
    Ours w/o $\mathcal{L}_\textit{depth}$ & 22.99 & 0.8170 & 0.117 \\
    Ours w/o $\mathcal{L}_\textit{motion}$ & 22.61 & 0.8027 & 0.137 \\
    Ours w/o rigidity & 22.73 & 0.8142 & 0.118 \\
    Ours & \textbf{23.65} & \textbf{0.8452} & \textbf{0.093} \\
    \bottomrule
    \end{tabular}
    }
\end{table}


\subsection{Ablation Study}
\tabref{ablation} analyzes the contribution of each loss quantitatively.

\begin{figure}[t]
\newlength\figwidthdepth
\setlength\figwidthdepth{0.245\linewidth}
\parbox[t]{\figwidthdepth}{\centering%
  \includegraphics[trim=230 20 100 60, clip=true, width=\figwidthdepth]{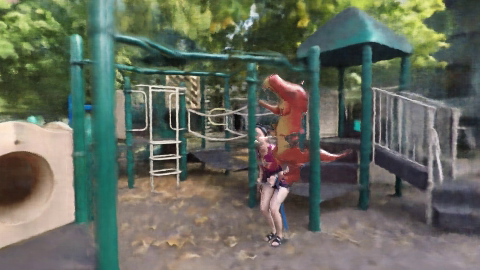}\\
  \small \emph{Without} depth order loss}%
\hfill
\parbox[t]{\figwidthdepth}{\centering%
  \includegraphics[trim=230 20 100 60, clip=true, width=\figwidthdepth]{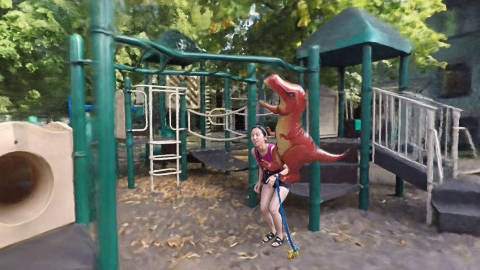}\\
  \small \emph{With} depth order loss}%
\hfill
\parbox[t]{\figwidthdepth}{\centering%
  \includegraphics[trim=270 35 60 45, clip=true, width=\figwidthdepth]{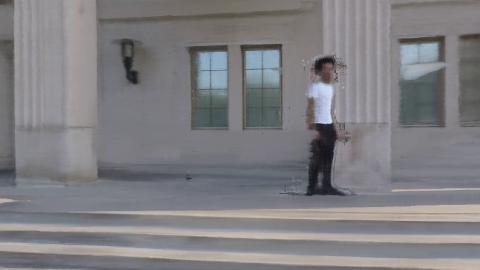}\\
  \small \emph{Without} motion regularization}%
\hfill
\parbox[t]{\figwidthdepth}{\centering%
  \includegraphics[trim=270 35 60 45, clip=true, width=\figwidthdepth]{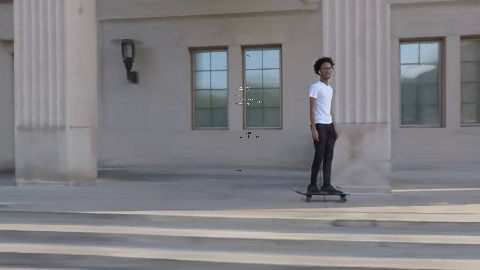}\\
  \small \emph{With} motion regularization}%
\vspace{\figcapmargin}
\caption{\textbf{Depth order loss and motion regularization.} Training with depth order loss ensures the correct relative depth of the dynamic object. Regularizing our scene flow prediction in dynamic NeRF can help handle videos with large object motion. 
}
\label{fig:ablation_depth}
\end{figure}

\topic{Depth order loss.}
For a complicated scene, we need additional supervision to learn the correct geometry. In \figref{ablation_depth} we study the effect of the depth order loss. 
Since the training objective is to minimize the image reconstruction loss on the input views, the network may learn a solution that correctly renders the given input video.
However, it may be a physically incorrect solution and produces artifacts at novel views.
With the help of the depth order loss $\mathcal{L}_\textit{depth}$, our dynamic NeRF model learns the correct relative depth and renders plausible content.

\topic{Motion regularization.}
Supervising scene flow prediction with the 2D optical flow is under-constrained.
We show in \figref{ablation_depth} that without a proper motion regularization, the synthesized results are blurry.
The scene flow may points to the wrong location.
By regularizing the scene flow with to be \emph{slow}, \emph{temporally and spatially smooth}, and \emph{consistent}, we obtain plausible results.

\topic{Rigidity regularization of the scene flow.}
The rigidity regularization helps with a more accurate scene flow prediction for the static region.
The dynamic NeRF is thus trained with a more accurate multi-view constraint.
We show in \figref{compare_Li} that the rigidity regularization is the key to a clean background.

\begin{figure}[t]
\newlength\figwidthfailure
\setlength\figwidthfailure{0.495\linewidth}
\parbox[t]{\figwidthfailure}{\centering%
  \fbox{\includegraphics[trim=0 0 0 0, clip=true, width=\figwidthfailure]{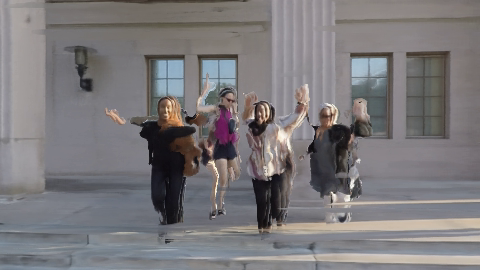}}\\
	\small Non-ridge deformation}%
\hfill
\parbox[t]{\figwidthfailure}{\centering%
  \fbox{\includegraphics[trim=0 0 0 0, clip=true, width=\figwidthfailure]{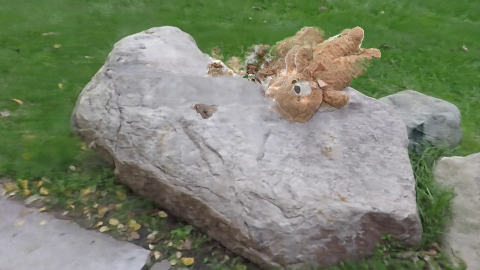}}\\
	\small Incorrect flow}%
\vspace{\figcapmargin}
\caption{\textbf{Failure cases.}
(\emph{Left}) Our method does not handle non-rigid deformation very well.
(\emph{Right}) Our dynamic NeRF heavily relies on the optical flow estimation and produces artifacts with inaccurate flow estimates.
}
\label{fig:failure}
\end{figure}

\subsection{Failure Cases}
Dynamic view synthesis remains a challenging problem. 
We show and explain several failure cases in \figref{failure}.
\section{Conclusions}
\label{sec:conclusions}
We have presented a new algorithm for dynamic view synthesis from a single monocular video. 
Our core technical contribution lies in scene flow based regularization for enforcing temporal consistency and alleviates the ambiguity when modeling a dynamic scene with only one observation at any given time.
We show that our proposed scene flow based 3D temporal consistency loss and the rigidity regularization of the scene flow prediction are the keys to better visual results.
We validate our design choices and compare favorably against the state of the arts. 

{\small
\bibliographystyle{ieee_fullname}
\bibliography{main}
}
\end{document}